\documentclass[conference]{IEEEtran}
\IEEEoverridecommandlockouts
\usepackage{cite}
\usepackage{caption}
\usepackage{enumerate}
\usepackage{amsmath}

\makeatletter
  \l@addto@macro\captionfont{\fontsize{7}{8}\selectfont}
\makeatother    
\usepackage{amsmath,amssymb,amsfonts}
\usepackage{algorithmic}
\usepackage{graphicx}
\usepackage{textcomp}
\usepackage{xcolor}
\usepackage{float}
\usepackage{array}
\usepackage[obeyspaces]{xurl}

\def\BibTeX{{\rm B\kern-.05em{\sc i\kern-.025em b}\kern-.08em
    T\kern-.1667em\lower.7ex\hbox{E}\kern-.125emX}}
\begin{document}

\title{Benchmarking of Different YOLO Models for CAPTCHAs Detection and Classification\\}

\author{\IEEEauthorblockN{Mikołaj Wysocki}
\IEEEauthorblockA{\textit{ITTI Sp.z o.o.} \\
Poznań, Poland \\
mikolaj.wysocki@itti.com.pl} 
\\
\IEEEauthorblockN{George Pantelis}
\IEEEauthorblockA{\textit{UBITECH} \\
Athens, Greece  \\
gpantelis@ubitech.eu}
\and
\IEEEauthorblockN{Henryk Gierszal}
\IEEEauthorblockA{\textit{Adam Mickiewicz University} \\
Poznań, Poland \\
gierszal@amu.edu.pl} 
\\
\IEEEauthorblockN{Sophia Karagiorgou}
\IEEEauthorblockA{\textit{UBITECH} \\
Athens, Greece \\
skaragiorgou@ubitech.eu}
\and 
\IEEEauthorblockN{Piotr Tyczka}
\IEEEauthorblockA{\textit{ITTI Sp.z o.o.} \\
Poznań, Poland \\
piotr.tyczka@itti.com.pl} 
\and

\and

}

\maketitle

\begin{abstract}
This paper provides an analysis and comparison of the YOLOv5, YOLOv8 and YOLOv10 models for webpage CAPTCHAs detection using the datasets collected from the web and darknet as well as synthetized data of webpages. The study examines the \textit{nano (n)}, \textit{small (s)}, and \textit{medium (m)} variants of YOLO architectures and use metrics such as \textit{Precision}, \textit{Recall}, \textit{F1} score, \textit{mAP@50} and inference speed to determine the real-life utility. Additionally, the possibility of tuning the trained model to detect new CAPTCHA patterns efficiently was examined as it is a crucial part of real-life applications. The image slicing method was proposed as a way to improve the metrics of detection on oversized input images which can be a common scenario in webpages analysis. Models in version \textit{nano} achieved the best results in terms of speed, while more complexed architectures scored better in terms of other metrics. 
\end{abstract}

\begin{IEEEkeywords}
Computer Vision, Object Detection, YOLO, CAPTCHA 
\end{IEEEkeywords}

\section{Introduction}

As Completely Automated Public Turing Test to Tell Computer and Humans Apart (CAPTCHA) \cite{vonAhn2003} is designed to protect data, copyrighted contents, servers, and services from bots. They can also be a blocking point for web-crawlers used to analyze illegal marketplaces and forums on the Dark Web, fake-new detectors, etc. CAPTCHAs are also used to prevent against spam and fraud, to check compliance, and to verify users. Classifying and pointing the position of a CAPTCHA code can be the first and crucial step of automatic CAPTCHA solving. Such a classifier can also serve as a tool for gathering the data about the most commonly used CAPTCHA types across the web. The development of the powerful machine learning algorithms made it possible to automatically solve different types of CAPTCHAs like text-based \cite{Kumar2004Aiips,  Bursztein2014, Jing2019} or image-based also called "reCAPTCHAs" \cite{Wang2020, Plesner2024}. Increasing number of automated CAPTCHA solvers (such as Optical Character Recognition (OCR) algorithms and Deep Learning Models) resulted in the development of new CAPTCHA types and schemes varying from simple puzzles and math, through audios and videos, to interactive and mini-games~\cite{Singh2014, Algwil2023}. With the increasing variant of CAPTCHAs the challenge might not be only to solve the CAPTCHA but also to determine its type and find its position on the webpage.  
Recent technological advances in computer vision and real-time detectors have opened the opportunity for accurate and precise tool for CAPTCHA recognition. 

There are many solutions used to the image recognition that are based on Machine Learning (ML) techniques like Neural Networks (NNs). One of the popular approaches involves the You Only Look Once  (YOLO) framework. The YOLO architecture \cite{JRed2015} as a Convolutional Neural Network (CNN) is well known for its accuracy and speed in computer vision tasks both while working with images and videos \cite{Jiang2022}.
Starting from the analysis of accessible datasets, through the possible methods of enlarging the training data, tackling the issue of diverse size of input images, and ending on the training tuning and then evaluating the YOLO-based models, this paper aims to explore the possibility of composing Artificial Intelligence (AI) driven CAPTCHA detector that provides a classification model to identify and CAPTCHA types and determine where CAPTCHA pattern is located in the webpage.

In this paper we present a complex method for collecting a ML training dataset and also performance results of ML models trained for automatic classification of CAPTCHA codes embedded in webpages. We describe how a dataset for the training process was collected in a few steps to increase the quantity and diversity of labeled specimens. We focus on YOLO models as mature solutions used in image recognition. Moreover, a smart technique to process images dived into horizontal slides is implemented in order to minimize a problem with rescaled images. All YOLO models are compared using a set of performance metrics.

The structure of the paper is as follows. In section II there is an overview of the related works. Section III provides an insight into the methodology. Subsection III.A describes the process of obtaining and preparing the dataset used to develop the CAPTCHA detector. In the following subsection III.B we propose the image slicing approach to overcome the problem of oversized input images. Subsection III.C gives an overview of training and evaluation details. The section III ends with the subsection III.D which provides the detailed description of performance metrics. Results obtained and their discussion are presented in Section IV. The final Section V contains conclusions.

\section{Related works}
Research to solve CAPTCHA codes automatically using ML techniques has become a challenge for years. There are even commercial services available in Internet where everyone can buy a packet of Application Programming Interface (API) requests to solve CAPTCHAs in webpages which are visited. That is why new CAPTCHA types like Google’s reCAPTCHAv2 or reCAPTCHAv3 systems are still being developed to counteract against bots that are feasible to provide CAPTCHA resolves in real time. To solve CAPTCHAs based on sequential information (like text of different lengths or with complicated features of characters) one can also concatenate a CNN with a Recurrent Neural Network (RNN) \cite{Shu_Xu} that is able to correlate dependencies in CAPTCHA patterns. Another ML-based solution is Generative Adversarial Networks (GANs) \cite{Zhang2020} that allow providing synthetic CAPTCHA images as similar as possible to the real ones. A promising technique is Capsule Networks \cite{Mocanu2022} that can detect the spatial relationships between different characters in the CAPTCHA. In \cite{Zahra2020} a customized CNN called Deep-CAPTCHA was developed to solve both numerical and alphanumerical CAPTCHAs, leading to cracking accuracy of 98.94\% and 98.31\%, respectively. The same approach based on a CNN named CapNet as well as VGG-19 and AlexNet deep CNN models were selected in \cite{Walia2023}. The CapNet solver achieved accuracy of 96.08\% (with slight differences for each of 5 digits in alphanumeric characters of CAPTCHA challenges) using advanced pre-processing techniques like noise reduction filtering, Grey-scaling, resizing, normalization, one-hot encoding, and image size reduction. Pre-trained YOLO v8 models for image segmentation and classification were used in \cite{Plesner2024} for three types of CAPTCHAs (and 13 classes) provided within the reCAPTCHAv2 system. According to the normalized confusion matrix top 5 accuracy was 99.5\%. Reinforcement Learning techniques can be used to provide automatically upgrades for CAPTCHA-solving algorithms.

\section{Methodology}
\subsection{Dataset preparation}

\begin{figure}[H]
    \centering
    \includegraphics[width=0.85\linewidth]{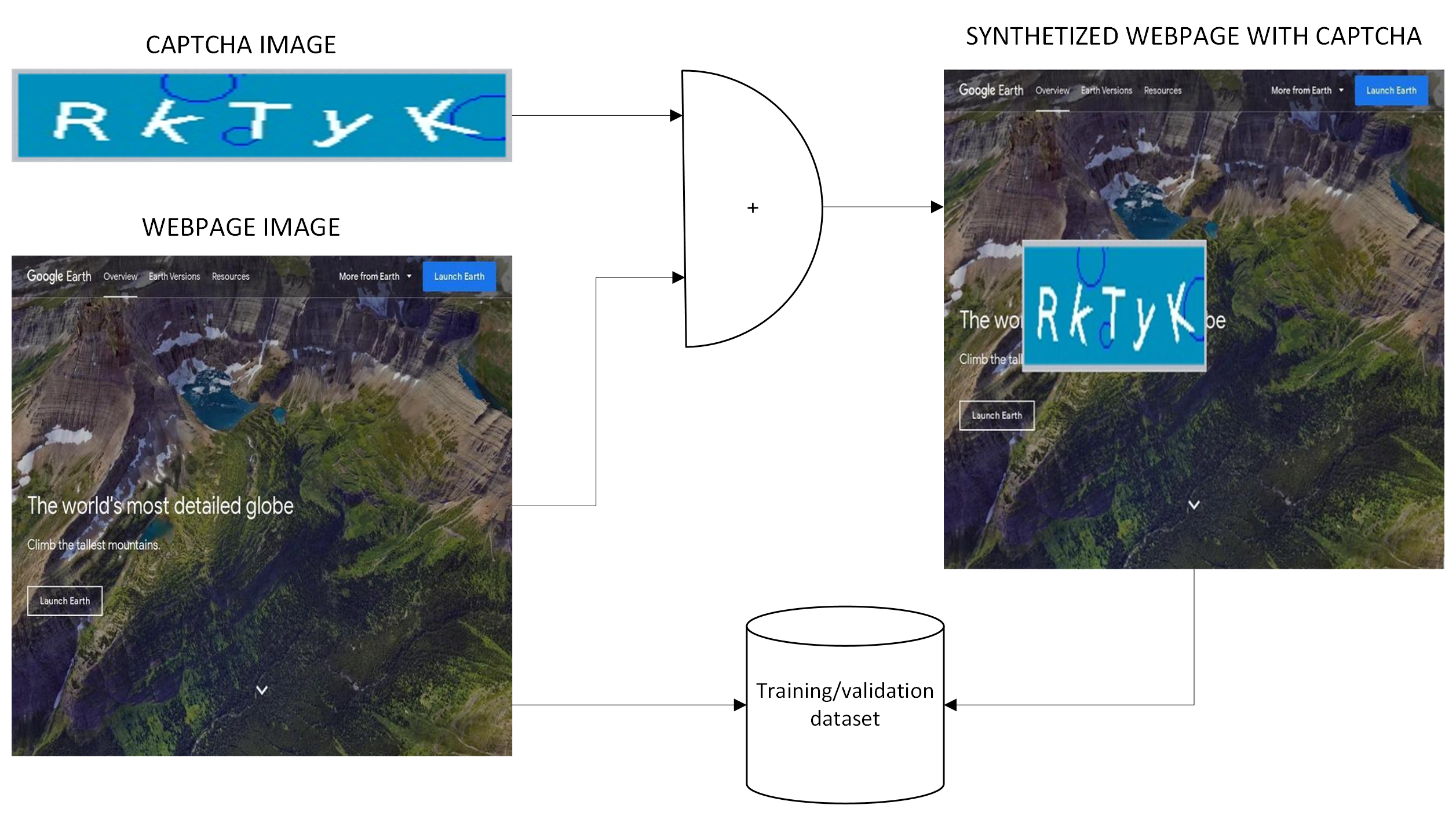}
    \caption{The CAPTCHA image is combined with the webpage image to obtain a synthetized image of the webpage protected by CAPTCHA. Both original and synthetized webpage images are then used in training of the neural network.}
    \label{fig:Captchasynth}
\end{figure}

\begin{figure}[H]
    \centering
    \includegraphics[width=1\linewidth]{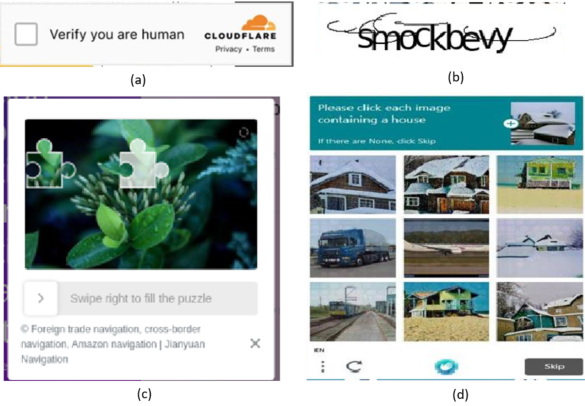}
    \caption{Four CAPTCHA images classes (a) button \cite{CloudCaptcha}, (b) text \cite{WikiCaptcha}, (c) puzzle \cite{PuzzleAmazon}, and (d) image \cite{Hcaptcha}.}
    \label{fig:Ctype}
\end{figure}

Starting point of the experiment was to collect the images of webpages protected by CAPTCHAs. Although collections of CAPTCHA images can be found on popular portals like \textit{Kaggle} or \textit{Robflow} those are mostly plain images of CAPTCHAs cropped from the whole website. To make the training dataset more diverse and increase the neural network confidence to distinguish CAPTCHAs from typical webpage elements which may be falsely classified as CAPTCHAs (like heading, images, adverts, etc.), it was decided to collect and combine three datasets: 
    \begin{enumerate}[(1)]
        \item
          Webpage images were obtained using the \textit{Selenium WebDriver} \cite{Selenium} with Python scripts as screenshots of ca. 10,000 most popular webpages based on the existing collection \cite{WebRank}. In addition to that subset, additional images were obtained from \textit{Kaggle} datasets \cite{Webpagedataset, WebpagedataSet2}.
        \item CAPTCHA images were gathered using the Selenium WebDriver with Python scripts as screenshots of: Amazon \cite{PuzzleAmazon}, Weibo \cite{Weibo}, Wikipedia \cite{WikiCaptcha}, and Shopee \cite{Shopee} webpages. Additional images were also obtained from \textit{Kaggel} \cite{Pcap, DarkTcap, CaptchaKaggle, CaptchaKaggle2, dataset, CaptchaKaggle3} and \textit{Robflow} datasets \cite{Creg, Rob2, Rob3}. Moreover, the \textit{PyCaptcha} \cite{PyCap} tool was used to generate text CAPTCHAs. Each CAPTCHA image was assigned to one of four classes: 
                \begin{enumerate}[a)]
                 \item Text CAPTCHAs
                  \item Puzzle CAPTCHAs
                    \item Image CAPTCHAs
                    \item Button CAPTCHAs
            \end{enumerate}
that have allowed preparing (3)
\item final training dataset.
    \end{enumerate}
Webpage images from the darkweb were collected using a data crawler alongside a Tor proxy \cite{torProxy} to enable access. Specifically, datasets were obtained from four different dark web sites to provide additional diversity to the training data.
To overcome the unavailability of various training sub-datasets of webpages protected by CAPTCHAs, images from the CAPTCHA dataset collected (dataset (2)) were randomly resized and randomly pasted onto the images from the webpage dataset (1), in order to obtain additional synthetic specimens of webpages protected by CAPTCHA. Such a solution not only significantly enlarged the training dataset (denoted in the following as (3)) but also simplified the process of labeling the data specimens (CAPTCHA type) by calculating the label based on CAPTCHA image width, height, and coordinate of leftmost corner on a plane of the webpage image.  Fig. \ref{fig:Captchasynth} depicts the process of combining those two datasets in order to use the resulting dataset for neural network training. 
The original webpage images without CAPTCHA (dataset (1)), used for synthetizing data specimens, were also added to the training dataset (3) with empty labels (no detection class - "unlabeled" in \ref{fig:datasetdist}) to reduce the number of false positive detections.
Fig. \ref{fig:Ctype} shows a single sample of each CAPTCHA class.

\begin{figure}[]
    \centering
    \includegraphics[width=0.8\linewidth]{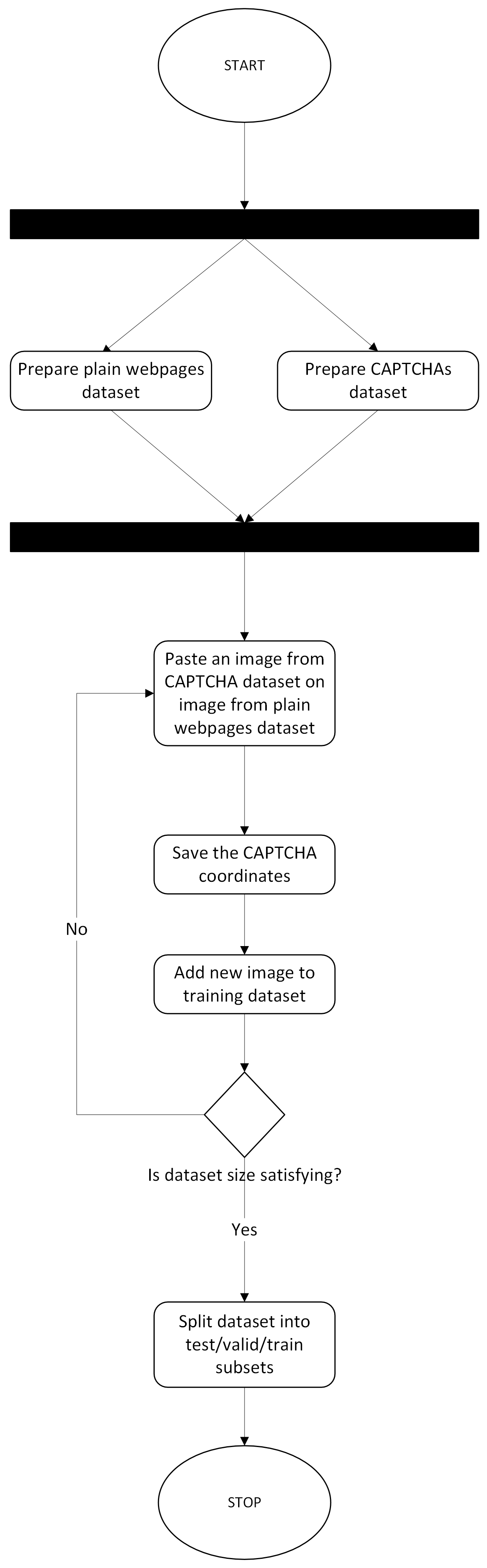}
    \caption{Preprocessing process flow diagram.}
    \label{fig:preprocessing}
\end{figure}

\begin{figure}[H]
    \centering
    \includegraphics[width=1\linewidth]{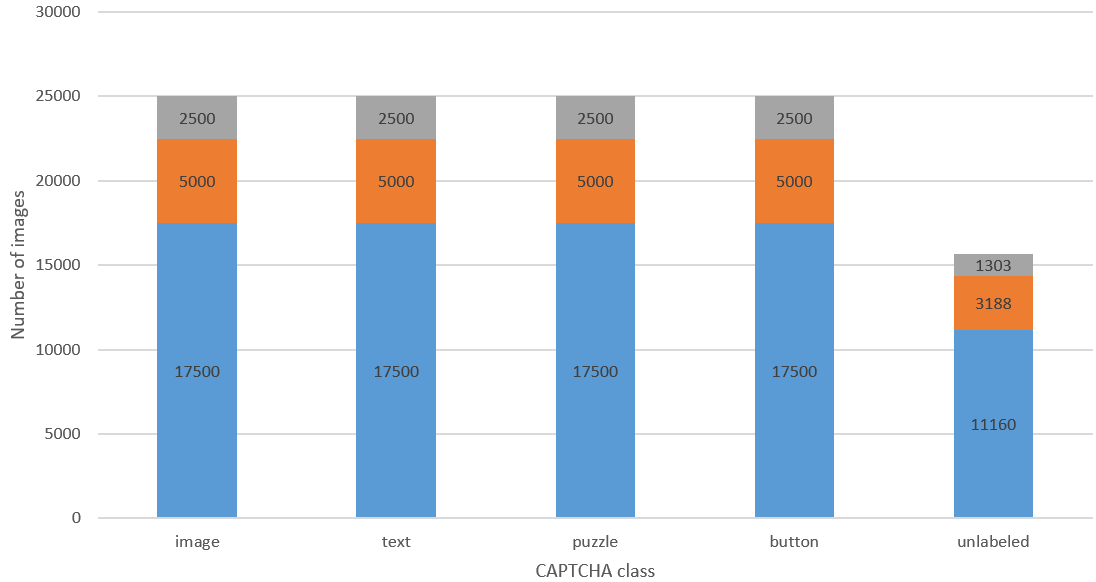}
    \caption{Distribution of classes in the dataset.}
    \label{fig:datasetdist}
\end{figure}

Fig. \ref{fig:preprocessing} presents a flow diagram containing all main steps of data preprocessing sequence. 
After collecting the datasets (plain webpage images (1) and CAPTCHA images (2)), the synthetized data are produced until the number of images in training dataset (3) meets the desired volume, which depends on available computational power (as the number of images in the training dataset increases, so does training time) and variety of images in both webpage (1) and CAPTCHA (2) datasets. Ideally the number of samples for each class (five classes in \ref{fig:datasetdist}) should be equal.
The number of samples used for the experiment, were 115,651. The last step of the preprocessing was to split the training dataset (3) between train, valid and test subsets as 70\%, 20\%, and 10\% of cardinality of the whole dataset. The exact distribution of the classes in the dataset is presented in Fig. \ref{fig:datasetdist}.
\subsection{Dividing the input image into slices}
A trained network would have a fixed size of input, but depending on the layout of the webpage and the method how the screenshot is taken, the input images will vary in size and resolution. The trained neural network will always resize the image to fit to the model input size, which in extreme cases may notably distort the image causing the significant drop in the performance of the model.

\begin{figure}[b]
    \centering
    \includegraphics[width=1\linewidth]{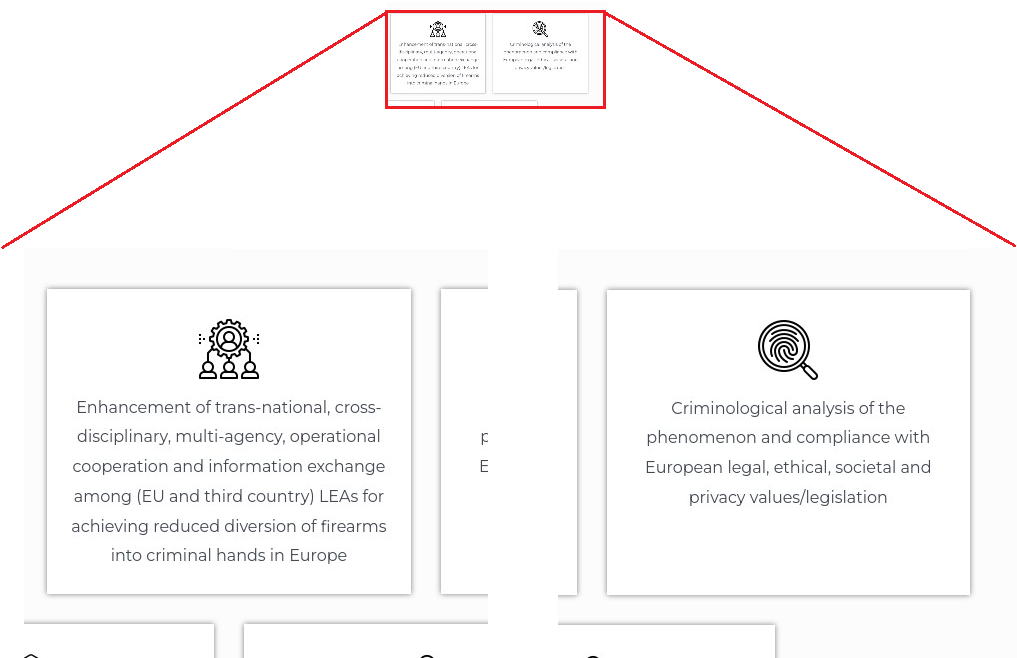}
    \caption{Rescaling the webpage image \cite{CeasefireWeb} degrades pixel resolution and  makes the text inside the red box unreadable. Slicing the original image prevents the information loss.}
    \label{fig:Ceasefire webpage split}
\end{figure}

\begin{figure}[b]
    \centering
    \includegraphics[width=1\linewidth]{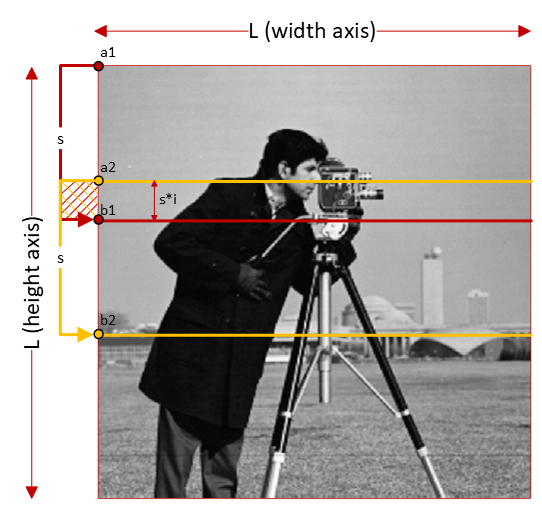}
    \caption{The first two pairs of the coordinates (\textit{a}, \textit{b}) calculated along the height axis produces red and yellow slices that overlap over \textit{s}\textit{*}\textit{i}  height.}
    \label{fig:cameramanSplit}
\end{figure}

To tackle this issue, an input image that exceeds a certain level of scale difference between current size and optimal one might be divided into smaller chunks of images (i.e. image slices). Each slice should contain the portion of the last slice to decrease the chance of having a desired classification object torn between neighbor slices. Fig. \ref{fig:Ceasefire webpage split} is a simple demonstration of rescaling problem and solution.
To calculate the starting and ending points of \textit{n} slices along the image width and height, the equation (\ref{eq:slice}) was used:
\begin{equation}    \label{eq:slice}
 (a_{n},b_{n})=
    \begin{cases}
        (0, s)&: n=1\\
        (L-s, L)&:b_{n-1}-s \times i+s>L\\
        (b_{n-1}-s \times i, a_{n}+s)&:otherwise     
    \end{cases}
\end{equation}
Where \textit{a} and \textit{b} are the starting and ending points of the slice along the axis of either input image width or height. \textit{L} is respectively input image width or height, \textit{s} is the slice size and \textit{i} is the percentage value of slices overlapping. Fig. \ref{fig:cameramanSplit} shows the equation parameters marked on an example of the image.

\subsection{Training and evaluation details}
In our study we decided to compare three YOLO models (versions) which are:
\begin{itemize}
    \item \textbf{YOLOv8:} \\
    Built on the foundations of the YOLOv5 and launched by Ultralitics in January 10\textsuperscript{th}, 2023 \cite{UltralyticsYOLO}. YOLOv8 introduced some key improvement; the most notable being anchor-free detection that increases the precision of the model comparing to its anchor-based ancestors \cite{anchorvsnon}, alongside with innovative feature named Spatial Pyramid Pooling Feature (SPPF) that improved model's performance in detecting objects of diverse scales \cite{YOLOv8eval}.      
    \item  \textbf{YOLOv5u:} \\
    YOLOv5u integrates the features introduced in YOLOv8 (anchor-free, objectness-free split head) to the architecture of YOLOv5 model as well as offers the big ammout of pre-trained models, improving the accuracy-speed tradeoff, and makes the model more flexible in adapting to different scenarios \cite{YOLOv5}. 
    \item \textbf{YOLOv10:} \\
    Built on the Ultralitics Python package \cite{UltralyticsYOLO} and launched in May 23\textsuperscript{rd}, 2024 by Tshingua University \cite{wang2024YOLOv10realtimeendtoendobject} aimed to find the balance between the performance and computational cost by eliminating Non-Maximum Suppression (NMS) to increase the postprocessing speed. Moreover YOLOv10 incorporates holistic efficient-accuracy driven model design strategy. Aditional optimization in the model architecture components aimed to reduce the computational overhead and improve overall model capabilities \cite{wang2024YOLOv10realtimeendtoendobject}.
\end{itemize}

Versions YOLO5u and YOLOv8 were chosen because of their availability and the ease of use. Both of them can be trained through Ultralitics hub \cite{UltralyticsYOLO} which might make them the first choice for the users. YOLOv10 was selected because it was the newest version of release for the time of conducting the experiments (in September 2024). 

We trained YOLO models based on the default parameters proposed by Ultralitics company. Each network was trained for 100 epochs. Moreover we conducted additional experiment in which we changed parameters of YOLOv10 model to see if we could obtain better results after enabling the cosine learning rate scheduler (cos\_lr) and increasing the number of training epochs. Default parameters for YOLO and modified parameters are presented in Table \ref{table:TrainingParameters}.

\begin{table}[H]
\centering
\caption{\textsc{Training parameters}}
\begin{tabular}{|>{\centering\arraybackslash}p{0.15\linewidth}|>{\centering\arraybackslash}p{0.1\linewidth}|>{\centering\arraybackslash}p{0.15\linewidth}|>{\centering\arraybackslash}p{0.2\linewidth}|>{\centering\arraybackslash}p{0.1\linewidth}|} \hline  
Setup & Learning rate & Optimizer  & Update of model: cos\_lr& Epochs \\ \hline  
Default YOLO & 0.01 & SGD & Disabled & 100 \\ \hline  
Modified YOLOv10 & 0.01 & SGD & Enabled & 234 \\ \hline 

\end{tabular}

\label{table:TrainingParameters}

\begin{minipage}{7cm}
\vspace{0.1cm}
\vspace{0.1cm}
\small Notes: SGD - Stochastic Gradient Descent
\end{minipage}
\end{table}

\begin{figure}[h]
    \centering
    \includegraphics[width=1\linewidth]{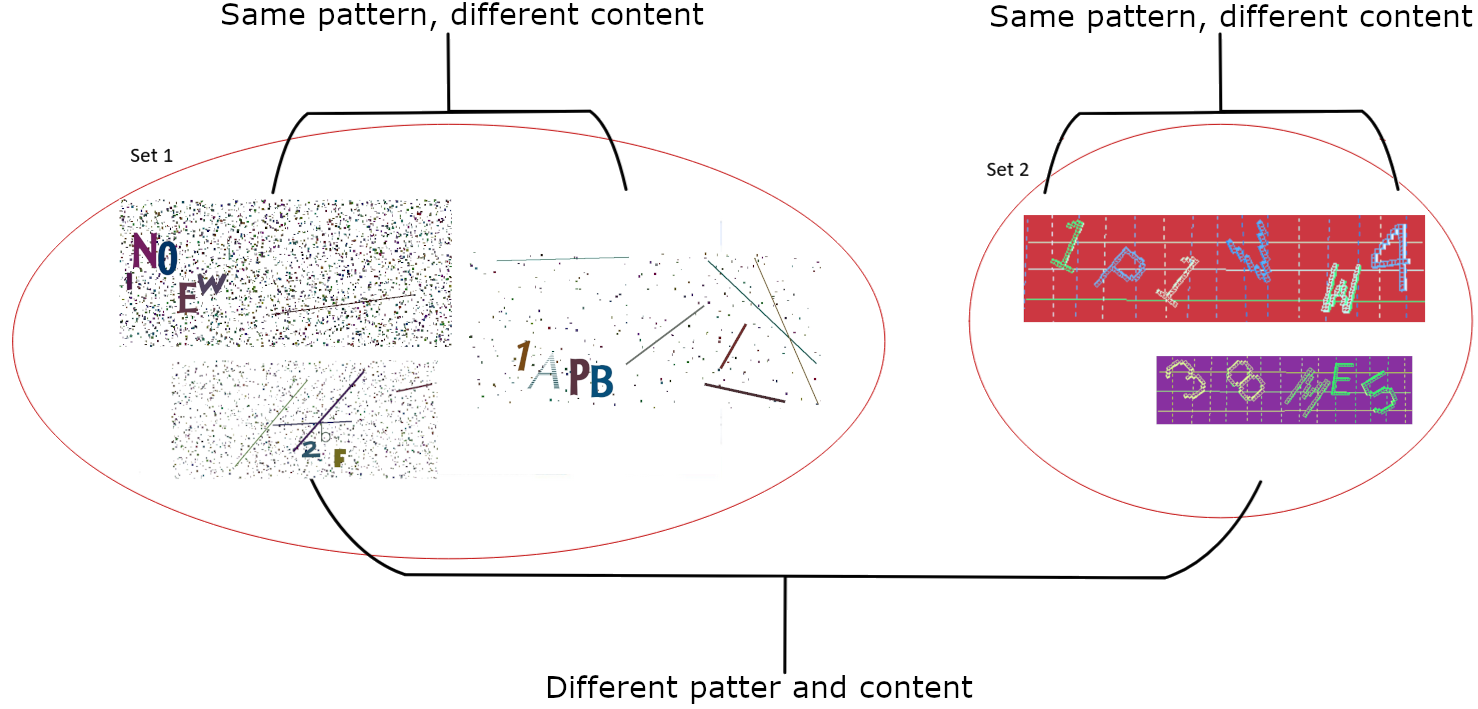}
    \caption{Differences in pattern and content between the text CAPTCHAs from Set 1 and Set 2}
    \label{fig:captchadiff}
\end{figure}

Following the training, the best model was evaluated by calculating the performance metrics for two test sets:
\\
\begin{itemize}
    \item \textbf{Set 1:} \\
    Part of the dataset depicted in Fig. 4. It contains the webpages with the same CAPTCHA pattern as in training and validation set but with a different content and background (webpage). A differences in the CAPTCHA pattern and content are presented in Fig. \ref{fig:captchadiff}.
   Set 1 was used to validate the generalization capabilities of the model, while working with the known CAPTCHA patterns.
    \item \textbf{Set 2:} \\
    It is an additional dataset with no synthetized data and the completely new CAPTCHA patters gathered from the web. This small set of 225 specimens was used to evaluate the capability of the trained neural network to classify CAPTCHAs of unknown pattern. Set 2 was also used to explore the possibility of tuning the neural network for classifying the new CAPTCHA patters with small amount of the data without the huge impact on the network performance.
    Because of the limited number of the source data and imbalance in the CAPTCHA types to be embedded in the webpages, classes’ instances in Set 2 were not equally numerous. The final class distribution in Set 2 was: text (79), image (74), puzzle (20), and button (52).
\end{itemize}
\subsection{Performance metrics}
\begin{enumerate}[a)]
    \item 

\textit{Precision}

\textit{Precision }measures the accuracy of the model’s positive predictions and is calculated via:
\begin{equation}
    \label{eq:precision}
    Precision=\frac{TP}{(TP+FP)}
\end{equation}

Where True Positive (TP) is the count of correct predictions and False Positive (FP) is the number of incorrect predictions of certain class. For the experiment, high \textit{Precision} means that the Neural Network (NN) model does not classify a fraction of webpage elements (text boxes, images, etc.) as CAPTCHAs and can correctly distinguish one CAPTCHA type from another. 

\item 
\textit{Recall}

\textit{Recall} (or \textit{sensitivity}) indicates the ability of the model to correctly detect and classify an object, and is defined by the formula below:
\begin{equation}
    \label{eq:Recall}
  Recall=\frac{TP}{(TP+FN)}
\end{equation}
Where False Negative (FN) stands for the number of the missed detections (class instances present on the image but classified as a different class or background). For the experiment, \textit{Recall }tells how good is the model in finding and properly classifying the CAPTCHAs on the webpage.

\item 
\textit{F1} score

\textit{F1} score is the combination of \textit{Precision} and \textit{Recall} to give the overall view on model performance. The equation that defines the \textit{F1 }score is  the following:
\begin{equation}
    \label{eq:f1}
    F1=2 \times \frac{Precision \times Recall}{Precision+Recall}
\end{equation}
Achieving the high \textit{F1} score can be interpreted as the model being able to correctly detect and classify the class instances, without missing or misclassifying the objects. For the CAPTCHA detector, \textit{F1} score indicates that the CAPTCHAs are properly localized and classified. \textit{F1} can also signal if the different CAPTCHA types are correctly distinguished from each other and if webpage elements are not misclassified as the CAPTCHAs. 

\item 
\textit{Mean Average Precision}

\textit{Mean Average Precision} (\textit{mAP}) is the metric that indicates the balance between \textit{Precision} and \textit{Recall}. It can be done by calculating the mean value of the area under the \textit{Precision-Recall} curve at different levels of \textit{Recall} for each detection class. \textit{mAP} is defined by formula (\ref{eq:mAP}):
\begin{equation}
    \label{eq:mAP}
    mAP=\frac{1}{n} \times \sum_{k=1}^{k=n}AP_{k}
\end{equation}
Where \textit{n} equals to number of classes and \textit{AP\textsubscript{k}} is average \textit{Precision} for class \textit{k}. For the CAPTCHA detector \textit{mAP} defines how good the task of detecting and classifying the CAPTCHAs is done across all of the detection classes. The metric used in experiments is mAP@50 which means that mAP is calculated considering all the detections that exceeds an intersection over union (\textit{IoU}) of 0.50.

\item 
\textit{Inference speed}

\textit{Inference speed} is a time that the trained model needed for a single detection. 
\end{enumerate}

\section{Experimental Results and discussion}
\subsection{Trained model performance}
\begin{figure}[b]
    \centering
    \includegraphics[width=1\linewidth]{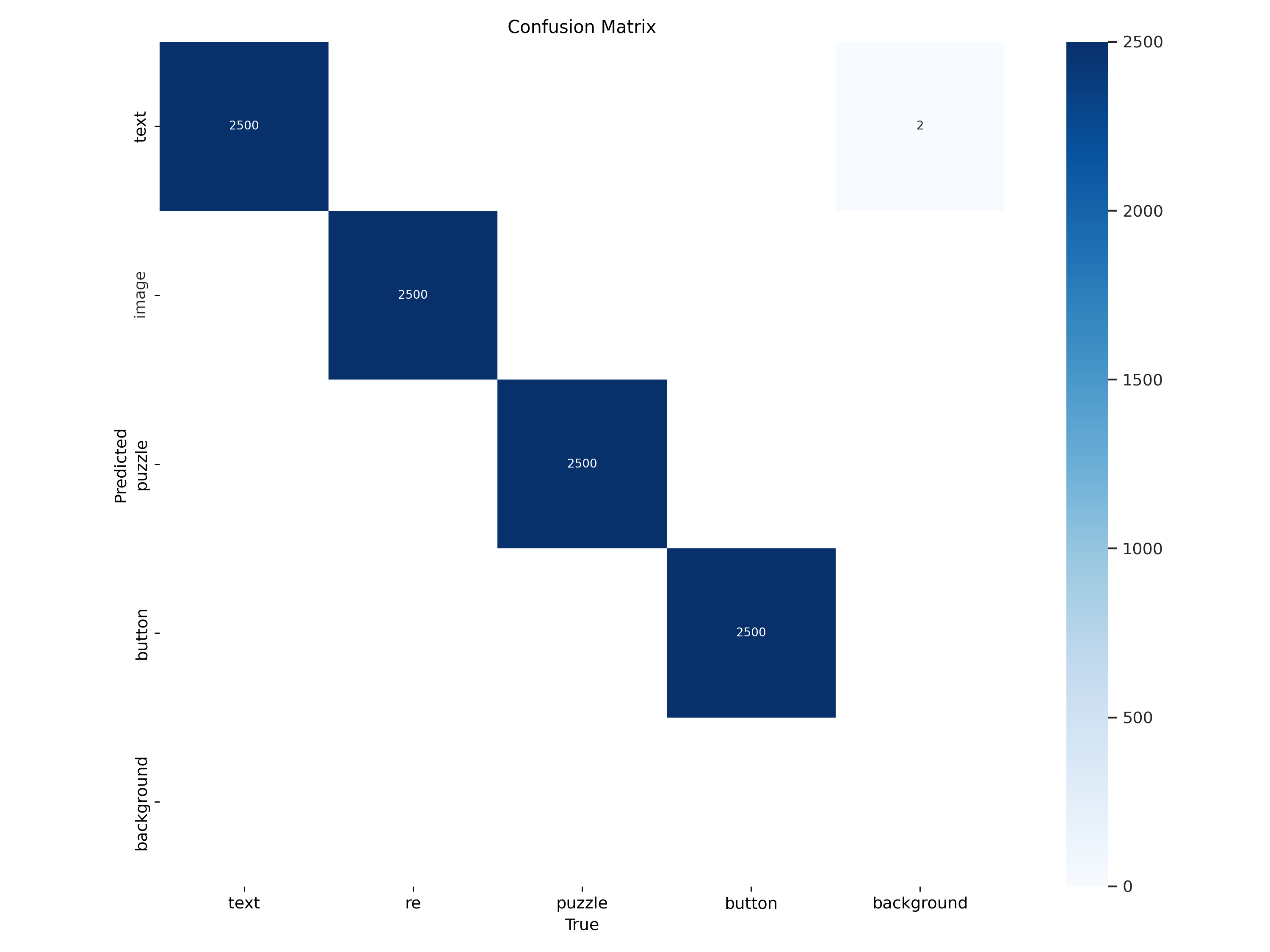}
    \caption{Confusion matrix (referring to YOLOv8m) as a result of the test on the Set 1. }
    \label{fig:cfmSet1}
\end{figure}

\begin{figure}[b]
    \centering
    \includegraphics[width=1\linewidth]{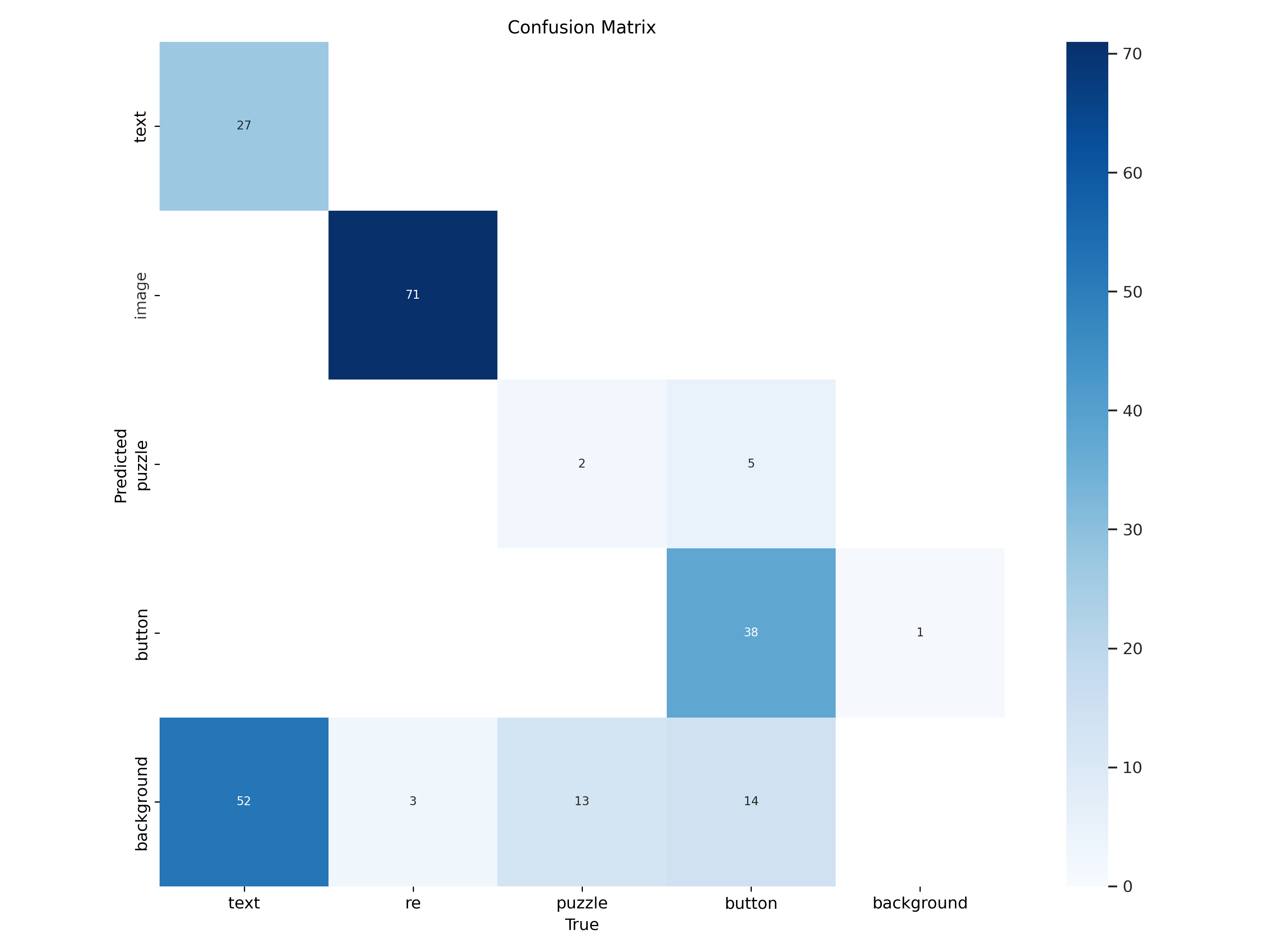}
    \caption{Confusion matrix (referring to YOLOv8m) as a result of the test on the Set 2. }
    \label{fig:cfmSet2}
\end{figure}

Table \ref{table:test resultsSet1} provides the metrics for each neural network model tested with Set~1 (the known CAPTCHA patterns). Letters \textit{‘n’}, \textit{‘s’}, and \textit{‘m’} stand for YOLO model type (nano, small, and medium), respectively. Additionally ‘YOLOv5un320’ represents the parameters for YOLOv5 model with an input size 320x320~[px], instead of 640x640~[px] (input size of other models).
Results are very similar and almost perfect. The differences in detection score between each pair of the models are marginal between 2 or 3 false detections. Such an outcome may be interpreted as a warning signal about overfitting. In such case, the results in Table \ref{table:test resultsSet2} as well as comparison of the confusion matrices in Fig. \ref{fig:cfmSet1} and Fig. \ref{fig:cfmSet2} were crucial for the experiment as they pointed the potential vulnerabilities. While the detection of the new CAPTCHA patterns for image class is good, there is a visible drop in performance of other classes (button, text, and puzzle). 

YOLOv5us and YOLOv10s models seems to be the best in terms of quality of detection, but they are slower than YOLOv8s and YOLOv5un. In terms of speed both YOLOv5 and YOLOv8 perform similar but YOLOv5un performed slightly better in tests on Set 2. Halving the input size of the YOLOv5un (to 320~px) doubled the detection speed without the performance drop.  

\begin{table*}[ht]
\centering
\caption{\textsc{Test results for Set~1 (known CAPTCHA patterns})}
\label{table:test resultsSet1}
\begin{tabular}{|>{\centering\arraybackslash}p{0.06\linewidth}|>{\centering\arraybackslash}p{0.05\linewidth}|c|c|c|c|c|c|c|>{\centering\arraybackslash}p{0.07\linewidth}|c|}
\hline
Model & YOLOv5 \newline un320& YOLOv5un &YOLOv5us& YOLOv8n & YOLOv8s & YOLOv8m  &YOLOv10n& YOLOv10s & YOLOv10s cos&YOLOv10m\\
\hline
\textit{Precision} & 0.998 & 0.998  &0.999& 0.999 & 0.999 & 0.999  &0.994& 0.999 & 0.998  &0.998\\
\hline
\textit{Recall} & 0.999 & 0.999  &0.999& 1 & 1 & 1  &0.994& 1 & 0.999  &0.999\\
\hline
\textit{F1} score & 0.998 & 0.998  &0.999& 0.999 & 0.999 & 0.999  &0.994& 0.999 & 0.998  &0.998\\
\hline
\textit{mAP} & 0.995 & 0.995  &0.995& 0.995 & 0.995 & 0.995  &0995& 0.995 & 0.995  &0.995\\
\hline
Inference speed& 0.4~ms & 0.8~ms

   &1.8~ms& 0.8~ms & 1~ms & 4.9~ms  &1~ms& 2.4~ms & 2.4~ms  &5.1~ms\\
\hline

\end{tabular}

\end{table*}

\begin{table*}[ht]
\centering
\caption{\textsc{Test results for Set~2 (unknown CAPTCHA patterns)}}
\label{table:test resultsSet2}
\begin{tabular}{|>{\centering\arraybackslash}p{0.06\linewidth}|>{\centering\arraybackslash}p{0.05\linewidth}|c|c|c|c|c|c|c|>{\centering\arraybackslash}p{0.07\linewidth}|c|}
\hline
Model & YOLOv5 \newline un320& YOLOv5un &YOLOv5us& YOLOv8n & YOLOv8s & YOLOv8m  &YOLOv10n& YOLOv10s & YOLOv10s cos&YOLOv10m\\
\hline
\textit{Precision} & 0.957 & 0.957  &0.959& 0.955 & 0.951 & 0.958  &0.942& 0.958 & 0.911  &0.957\\
\hline
\textit{Recall} & 0.596 & 0.596  &0.618& 0.569 & 0.6 & 0.613  &0.573& 0.609 & 0.636  &0.591\\
\hline
\textit{F1} score & 0.735 & 0.735  &0.752& 0.713 & 0.736 & 0.748  &0.713& 0.745 & 0.749  &0.731\\
\hline
\textit{mAP} & 0.653 & 0.653  &0.672& 0.644 & 0.632 & 0.665  &0.664& 0.670 & 0.665  &0.665\\
\hline

\end{tabular}

\end{table*}


\subsection{Image slicing test}
To check if proposed input slicing method can increase the performance of the model, the images from the Set 2 with a width or height higher or equal to 1920~px (three times neural network model input size) have been sliced using the formula (equation (\ref{eq:slice})) from Section~III.B.

30 images that exceed 1920~px threshold have been obtained from Set 2. After slicing the images, we tested the response of the neural network based on model YOLOv8m for sliced and unsliced images, calculating TP, FP, and FN metrics. Results are presented in Table \ref{table:slicingresults}. It can be observed that the image slicing can be a response to the problem of oversized input images, but it is important to keep in mind that such a method can have a huge impact on the model speed. The detection has to be done for each image slice. There is also a small risk of creating false positive detections. 
\begin{table}[H]
\caption{\textsc{Detection results on sliced images}}
\label{table:slicingresults}
\centering
\begin{tabular}{|c|c|c|c|}
\hline
Mode (sliced/unsliced) & TP & FN & FP \\
\hline
Unsliced & 20 & 10 & 0 \\
\hline
Sliced & 25 & 5 & 1 \\
\hline

\end{tabular}

\end{table}
\subsection{Tuning the network}
Analysis of the result tables (Table~\ref{table:test resultsSet1} and Table~\ref{table:test resultsSet2}) and confusion matrices (Fig.~\ref{fig:cfmSet1} and Fig.~\ref{fig:cfmSet2}) with the succeeding reanalysis of the training dataset pinpointed to the problem of puzzle CAPTCHA instance representation being not enough diverse (this type is used rather seldom) and also, as expected, some of the new patters of the text CAPTCHAs were not recognized by the trained neural network (cf. Fig.~\ref{fig:cfmSet2}). 

\begin{table*}[ht]
\centering
\caption{Test results of the retrained models}
\label{tab:retrainedresults}
\begin{tabular}{|c|c|c|c|c|}
\hline
Model/testing set & YOLOv5un320 Set~1 & YOLOv8m Set~1 & YOLOv5un320 Set~2 & YOLOv8m Set2 \\
\hline
\textit{Precision} & 0.999  & 0.999 & 0.953 & 0.963 \\
\hline
\textit{Recall} & 0.998  & 1 & 0.653 & 0.741 \\
\hline
\textit{Precision} for text class & 0.998  & 0.996 & 1 & 1 \\
\hline
\textit{Recall} for text class & 0.993  & 1 & 0.434 & 0.583 \\
\hline

\end{tabular}

\end{table*}

\begin{table*}[ht]
\centering
\caption{Test results without the retraining}
\label{tab:noretrainingresults}
\begin{tabular}{|c|c|c|c|c|}
\hline
Model/testing set & YOLOv5un320 Set~1 & YOLOv8m Set~1 & YOLOv5un320 Set~2 & YOLOv8m Set2 \\
\hline
\textit{Precision} & 0.998 & 0.999 & 0.957 & 0.958 \\
\hline
\textit{Recall} & 0.999 & 1 & 0.596 & 0.613 \\
\hline
\textit{Precision} for text class & 0.998 & 0.999 & 0.818 & 1 \\
\hline
\textit{Recall} for text class & 0.993 & 1 & 0.217 & 0.341 \\
\hline

\end{tabular}

\end{table*}

As a possible solution, the additional training session was conducted to tune the network for analyzing new instances of text CAPTCHAs. The portion of 1,000 images of the new instances of text CAPTCHAs was mixed with the random images from the previous training dataset (cf. Section~III.A), resulting in a new set of 34,304 images, divided into the training and validation subsets (26,154/8,150). The number of images from the previous training included in this retraining dataset was selected as a result of experiment. Training on such a dataset made it possible to classify the new text CAPTCHA patterns without the significant decay in recognizing the patterns learned before. Training was conducted for 50~epochs. Table \ref{tab:retrainedresults} and Table \ref{tab:noretrainingresults} show the network performance metrics for the retrained neural network (Table \ref{tab:retrainedresults}) and the neural network without this additional tuning (Table \ref{tab:noretrainingresults}), respectively.

Retraining of the network not only increased the number of true positive detections of the text class in Set 2 for both neural networks, but also eliminated the false positive detections of the text class for YOLOv5un320. A small drop of metrics can be observed for detection on Set~1 but it is a small cost of learning new text CAPTCHA patterns.

\subsection{Performance comparison of the YOLO models}
If the inference speed is omitted as a less important metric related to the hardware computing power, the ML models developed can be compared focusing on the other performance metrics listed in Section~III.D and evaluated in Table~\ref{table:test resultsSet2}. For this purpose a weighted arithmetic mean has been calculated using the following weights assigned to the metrics: F1: 50\%, mAP: 25\%, Precision: 12.5\%, and Recall: 12.5\%. According to this value a ranking of the models, starting from the best one, is: YOLOv5us, YOLOv8m, YOLOv10s, YOLOv10scos, YOLOv10m, YOLOv5un320, YOLOv5un, YOLOv8s, YOLOv10n, and YOLOv8n. When two models were retrained (cf. Section~IV.C), an average improvement for both Sets was 4\% for Recall and 44\% for Recall (cf. Table~\ref{tab:retrainedresults} and Table~\ref{tab:noretrainingresults}).

\section{Conclusion and steps ahead}
In this study we examined the capability and performance of different YOLO models in the task of detecting and classifying four distinct CAPTCHA types. This was motivated by the need to enhance the capabilities of the darknet web-crawler with the tool for  detecting and solving text-CAPTCHAs to be able to explore the darknet more freely in search for the information of interest. Our analysis involved various metrics such as \textit{Precision}, \textit{Recall}, \textit{F1-score}, \textit{mAP}, and inference time.

To overcome the issue of not enough good quality data, we created the synthetized training data of webpages protected by CAPTCHA code. We tested the trained neural networks both on synthetized data and real-life scenario that included new CAPTCHA patterns. While the detection of the image CAPTCHA class was efficient for both synthetized and real-life specimens, the metrics for other classes were worse. The poor results of real-life detection resulted from the high variety of different puzzle and text CAPTCHA patterns. To overcome this issue an additional training with new text CAPTCHA patterns present in the real-life dataset has been done to increase the confidence of the model in recognizing CAPTCHA patterns.

Conducted experiment proved that it is possible to tune the network for recognizing new CAPTCHA patterns using the small amount of data (in the experiment 1,000 synthetized webpage images) but it is important to keep old CAPTCHA patterns in the training and validation sets to minimize the risk of learning new patterns while forgetting the old ones.

The proposed slicing solution for dealing with oversized images tends to be working efficiently and can increase the number of true positive detections in cost of slower response. As a way of enhancing this method, the real-life websites can be examined for the most common location of the CAPTCHAs on the webpage (top, middle, and bottom). With statistics like these above, sliced images could be analyzed starting from the slice where there is the highest probability of the CAPTCHA occurring.

Different YOLO models analyzed differ mostly in better performance metrics in cost of speed. The final model to be chosen as recommended one depends on the application. In the real-time environment YOLOv5un and YOLOv8n outscore other models because of their speed. On the other hand, in the application where \textit{Precision} and \textit{Recall} metrics are more valid than speed, other models such as YOLOv8m and YOLOv10s will work better.

It is possible to train the network for recognizing and localizing the CAPTCHA types on the webpage using YOLO models, but the diverse dataset is crucial for obtaining good results. While it is possible to train the model on the synthetized data as well as to work with real-life specimens, performance will mostly depend on diversity of CAPTCHA patterns used for training. That is why an important part of building the CAPTCHA classification and recognition model should be a constant collecting of the CAPTCHA data for future network tuning.

\section*{Acknowledgment}
The research leading to these results has received funding from the European Union’s Horizon Europe research and innovation programme under grant agreement No.~101073876 - CEASEFIRE project \cite{CeasefireWeb}.
  
\vspace{4pt}
\bibliographystyle{IEEEtran}
\bibliography{references}
\smallskip

\end{document}